\documentclass[journal,10pt]{IEEEtran}

\usepackage[cmex10]{amsmath}
\usepackage[utf8]{inputenc}
\usepackage[pdftex]{graphicx}
\usepackage{epstopdf}
\usepackage{array}
\usepackage{floatrow}
\usepackage{caption}
\usepackage{subcaption}
\usepackage{url}
\usepackage{balance}
\usepackage{enumerate} 
\usepackage{amsbsy}
\usepackage[normalem]{ulem}
\usepackage{textcomp}
\usepackage{siunitx}
\usepackage{balance}
\usepackage{placeins}
\usepackage{tabularx, ragged2e}
\usepackage{listings}
\usepackage[dvipsnames]{xcolor}
\usepackage{svg}
\usepackage{amssymb}
\usepackage{stfloats}

\usepackage{booktabs}
\usepackage{multirow}
\usepackage{adjustbox}
\usepackage{tabularx}
\usepackage{pifont}

\usepackage{hyperref}

\usepackage[compact]{titlesec}
\titlespacing*{\subsection}{0pt}{0.8ex plus .2ex}{0.3ex}

\usepackage{cleveref}

\crefname{section}{Sec.}{Secs.}
\Crefname{section}{Sec.}{Secs.}
\crefname{subsection}{Sec.}{Secs.}
\Crefname{subsection}{Sec.}{Secs.}
\crefname{subsubsection}{Sec.}{Secs.}
\Crefname{subsubsection}{Sec.}{Secs.}
\crefname{figure}{Fig.}{Figs.}
\Crefname{figure}{Fig.}{Figs.}
\crefname{table}{Tab.}{Tabs.}
\Crefname{table}{Tab.}{Tabs.}
\crefname{equation}{Eq.}{Eqs.}
\Crefname{equation}{Eq.}{Eqs.}
\crefname{algorithm}{Alg.}{Algs.}
\Crefname{algorithm}{Alg.}{Algs.}
\crefname{appendix}{App.}{Apps.}
\Crefname{appendix}{App.}{Apps.}

\def\BibTeX{{\rm B\kern-.05em{\sc i\kern-.025em b}\kern-.08em
    T\kern-.1667em\lower.7ex\hbox{E}\kern-.125emX}}
    
\usepackage{xspace}

\DeclareTextFontCommand{\textcomputer}{\fontfamily{cmr}\selectfont}

\newcommand{\nicepar}[1]{\smallskip \noindent \textbf{#1}}

\newcommand{\qmarks}[1]{``#1''}

\newcommand{\methname}{CAMNet}
\newcommand{\extendedmethname}{\textbf{C}ooperative \textbf{A}wareness \textbf{M}essage-based Graph Neural \textbf{Net}work}

\newcommand{\tightEq}{\setlength\abovedisplayskip{8pt}\setlength\belowdisplayskip{12pt}}

\begin{document}

\bstctlcite{IEEEexample:BSTcontrol}

\title{\methname{}: Leveraging Cooperative Awareness Messages for Vehicle Trajectory Prediction}

\author{\IEEEauthorblockN{Mattia~Grasselli, Angelo~Porrello
        and Carlo~Augusto~Grazia}\\
\IEEEauthorblockA{\textit{Department of Engineering ``Enzo Ferrari''}
\textit{University of Modena and Reggio Emilia}\\
\{name.surname\}@unimore.it}\thanks{The work of Carlo Augusto Grazia was carried out within the MOST – Sustainable Mobility National Research Center and received funding from the European Union Next-GenerationEU (PIANO NAZIONALE DI RIPRESA E RESILIENZA (PNRR) – MISSIONE 4 COMPONENTE 2, INVESTIMENTO 1.4 – D.D. 1033 17/06/2022, CN00000023). Angelo Porrello was financially supported by the Italian Ministry for University and Research – through the ECOSISTER ECS 00000033 CUP E93C22001100001 project – and the European Commission under the Next Generation EU programme PNRR.}
}

\maketitle
\thispagestyle{empty}  %

\begin{abstract}
\looseness=-1
Autonomous driving remains a challenging task, particularly due to safety concerns. Modern vehicles are typically equipped with expensive sensors such as LiDAR, cameras, and radars to reduce the risk of accidents. However, these sensors face inherent limitations: their field of view and line of sight can be obstructed by other vehicles, thereby reducing situational awareness. In this context, vehicle-to-vehicle communication plays a crucial role, as it enables cars to share information and remain aware of each other even when sensors are occluded. One way to achieve this is through the use of Cooperative Awareness Messages (CAMs).

\looseness=-1
In this paper, we investigate the use of CAM data for vehicle trajectory prediction. Specifically, we design and train a neural network, \extendedmethname{} (\methname{}), on a widely used motion forecasting dataset. We then evaluate the model on a second dataset that we created from scratch using Cooperative Awareness Messages, in order to assess whether this type of data can be effectively exploited. Our approach demonstrates promising results, showing that CAMs can indeed support vehicle trajectory prediction. At the same time, we discuss several limitations of the approach, which highlight opportunities for future research.
\end{abstract}


\section{Introduction}
\label{sec_intro}
\looseness=-1
The idea of creating objects capable of acting autonomously has long captured the human imagination. Recent advancements in artificial intelligence have brought this vision closer to reality, enabling machines to perform increasingly complex tasks without human intervention. A notable example is autonomous driving, where substantial development has been made in perception, planning, and control systems \cite{Survey_autonomous_driving}.

\looseness=-1
However, mass production of autonomous vehicles will only become possible when sufficient safety is verified. A key feature in this context is the prediction of future states of surrounding vehicles in a manner comparable to human reasoning. Despite extensive research, accurate trajectory prediction continues to be an open challenge due to the diversity of traffic behaviors, complex agent interactions, and the inherent uncertainty in sensor data \cite{Survey_autonomous_driving}.

\looseness=-1
\indent One possible way to improve the safety for autonomous vehicles is to leverage the increasing availability of inter-vehicular communication data provided by modern Intelligent Transportation Systems (ITS). These systems enable vehicles to exchange real-time information with one another and with infrastructure, enhancing overall situational awareness in dynamic traffic environments. One type of data exchanged in this context is the  Cooperative Awareness Message (CAM). CAMs are designed to enable vehicles to maintain awareness of each other and "to support cooperative performance of vehicles using the road network" \cite{ETSI-v141}. 

\looseness=-1
\indent This paper investigates whether vehicle trajectories could be accurately predicted using only CAMs. Current autonomous driving systems rely on onboard sensors, such as LiDAR, cameras, and radars, to perceive the environment. However, these sensors are inherently limited by their field of view and line-of-sight constraints. As a result, vehicles that are occluded or outside the sensor range may go undetected, leading to incomplete awareness and, thus, suboptimal decision-making. CAM data can offer a complementary source of perception. If trajectory prediction based on CAM data proves feasible, it could significantly enhance autonomous driving performance by extending awareness beyond the physical limitations of onboard sensors.

\looseness=-1
Our contributions can be summarized as follows:
\begin{itemize}
    \item Construction of a dataset based solely on Cooperative Awareness Messages extracted from various Road-Side Units being part of the Modena Automotive Smart Area (MASA)~\footnote{MASA: \href{https://www.automotivesmartarea.it/?lang=en}{https://www.automotivesmartarea.it/}}, a living research lab in Modena, Italy.
    \item Analysis of the designed neural network on two datasets to evaluate whether CAM data are suitable for predicting vehicles' trajectories.
\end{itemize}

\looseness=-1
\indent The paper is organized as follows. \cref{sec_related} overviews related works. \cref{sec_problem_form} describes the problem formulation, while \cref{sec_dataset} presents the datasets utilized. \cref{sec_nn} analyzes the designed neural network, while \cref{sec_experiments} explores the experiments conducted and the results obtained. \cref{sec_conclusions} concludes the article. 

\section{Related Works}
\label{sec_related}
\looseness=-1
\nicepar{Motion Forecasting Datasets.} In the last few years, various large-scale datasets have been proposed and made public for training neural networks for predicting vehicles' trajectories. They generally vary depending on the data they exploit, the number of unique scenarios, and the total hours. One of the leading datasets for motion forecasting has been Argoverse \cite{chang2019argoverse}, and its successor Argoverse 2.0 \cite{wilson2023argoverse}, which became widely recognized as the first large-scale dataset to evaluate the impact of high-definition maps (HD-maps) on motion forecasting. INTERACTIONS \cite{interactiondataset}, instead, underlined the necessity of creating datasets that considered various driving scenarios from different countries. However, many others have been published during the years, such as Waymo Open Motion \cite{ettinger2021large} and Lyft Level 5 \cite{houston2021one}.

\looseness=-1
Due to the strong interest in this challenge, it has also occurred that famous datasets initially conceived for different tasks, such as perception, have been adapted for this challenge. That is the case, for instance, of NuScenes \cite{caesar2020nuscenes}. Notable is also V2X-Seq \cite{yu2023v2x}, which is a sequential dataset within the DAIR-V2X \cite{yu2022dair} family, that analyzes the insertion of infrastructure information, such as videos gathered using cameras, for vehicle-infrastructure cooperative trajectory forecasting. Unlike all the datasets previously analyzed, ours, which cannot be considered large-scale due to its small dimensions, focuses on how motion forecasting can be performed using data already transmitted by vehicles. To the best of our knowledge, no one has ever tried to produce datasets using Cooperative Awareness Messages. If feasible, it could enable the usage of this type of data to predict vehicle trajectories.

\looseness=-1
\nicepar{Trajectory Prediction Methods}. Over the years, many strategies have been employed for predicting trajectories of agents. One of the most basic, but effective, methods is the Constant Velocity Model (CVM), which assumes the vehicle will continue to move in the same direction and velocity as observed from the last two time steps \cite{scholler2020constant}. Even though such an approach is very simplistic, it has been shown to produce decent results, and it has become a standard comparison metric. This aspect -- namely, predicting trajectories starting from only a few observed points -- has also been explored in the context of complex neural networks specifically trained to transfer knowledge from models using a higher number of observations to those operating with fewer ones \cite{monti2022many}.

\looseness=-1
With the advancements in deep learning, many neural networks have also been proposed. For instance, LaneGCN \cite{liang2020learning} was the first neural network to process HD-maps in a vectorized (graph) form by exploiting a revised version of the Graph Convolutional Network proposed by \textit{Kipf and Welling} \cite{kipf2017semisupervisedclassificationgraphconvolutional}. Another neural network, called Forecast-MAE \cite{cheng2023forecast}, adapts the masked autoencoder (MAE) \cite{he2022masked} initially proposed in the Computer Vision field, for motion forecasting. Unlike prior neural network approaches, ours combines the VAE, RNN, and GNN, which is a solution not commonly found in the vehicle motion forecasting field.

\section{Problem formulation}
\label{sec_problem_form}
\looseness=-1
The task of trajectory prediction involves forecasting future positions of agents (e.g., vehicles) given their current and past observation states. The problem can be mathematically defined as follows: let \text{$\textbf{p}_i^t=(x_i^t,y_i^t,v_i^t,\theta_i^t,...)\in\mathbb{R}^m$} describe a generic actor at time-step \text{$t$} where \text{$\textbf{x}_i^t=(x_i^t,y_i^t)$} denote the actor's position, \text{$v_i^t$} indicate its velocity and \text{$\theta_i^t$} represent its orientation, while \text{$\ldots$} describes the remaining actor's features.
\looseness=-1
\indent The goal of agent motion forecasting is to design a model capable of predicting the future states \text{$\mathcal{Y}_i=(\mathbf{p}{^{t+1}_i},..,\mathbf{p}{^{t+T_{pred}}_i})$} of agent $i$, given its past observation states \text{$\mathcal{X}_i=(\mathbf{p}{^{t-T_{obs}}_i},..,\mathbf{p}{^{t}_i})$}, and, eventually, also the ones from its neighboring agents \text{$\{\mathcal{X}_j: j \neq i\}$} \cite{scholler2020constant}.

\looseness=-1
\indent During the study, Cooperative Awareness Messages will be exploited to predict vehicles' trajectories. CAMs contain extensive useful data, with particular relevance to this study being the vehicle's position, speed, and heading. Following the ETSI Standard \cite{ETSI-v141}, the generation of Cooperative Awareness Messages always periodically occurs within a second (1Hz); however, if a vehicle undergoes a significant change in position, speed, or heading compared to its last transmitted CAM, a new one is generated, with a maximum frequency of 10 Hz. In \cref{tab:CAM_triggers} are described in detail the CAM generation triggers.

\begin{table}[t]
    \centering
    \renewcommand{\arraystretch}{1.25}
    \resizebox{\textwidth}{!}{%
    \begin{tabular}{l c p{4.5cm}}
    \toprule
    \textbf{Trigger} & \textbf{Formula} & \textbf{Description} \\
    \midrule
    \textbf{Time} & \text{$(t_{\operatorname{current}}-t_{lastCAM})>1s$} & Time elapsed with respect to the last transmitted CAM is greater than 1 second. \\
    \textbf{Position} & \text{$||\textbf{x}_{\operatorname{current}}-\textbf{x}_{\operatorname{lastCAM}}||_2>4m$} & Vehicle has moved more than 4 meters with respect to its last CAM. \\
    \textbf{Heading} & \text{$|\theta_{\operatorname{current}}-\theta_{\operatorname{lastCAM}}|>4$°} & Vehicle has changed its heading more than 4° relative to its last CAM. \\ 
    \textbf{Speed} & \text{$|v_{\operatorname{current}}-v_{\operatorname{lastCAM}}|>0.5m/s$} & Vehicle has changed its speed more than 0.5 m/s relative to its last CAM. \\ 
    \bottomrule
    \end{tabular}
    }
    \caption{CAM generation triggers.}
    \label{tab:CAM_triggers}
\end{table}

\section{Datasets}
\label{sec_dataset}
\looseness=-1
Two datasets are used in this study. The first one is Argoverse 2 Motion Forecasting \cite{wilson2023argoverse}, which is a widely used dataset in motion forecasting research. It is employed for training and evaluating the designed neural network to enable a fair comparison with various competitors. The second dataset was created from scratch using CAM data collected over approximately one month in Modena, Italy, through 11 Road-Side Units (RSUs) deployed in the MASA living lab.

\looseness=-1
\subsection{Argoverse 2 Motion Forecasting Dataset}
Argoverse 2 Motion Forecasting Dataset is composed of $250000$ non-overlapping scenarios mined for interesting and challenging interactions among vehicles \cite{wilson2023argoverse}. The scenarios have been gathered from six distinct cities in the United States of America: Austin, Detroit, Miami, Palo Alto, Pittsburgh, and Washington D.C, and they are 11 seconds long, where the first 5 seconds denote the observation window, while the following 6 denote the forecasting horizon. Each scenario includes: a High-Definition Map (HD-Map) which provides the context information, and the trajectory data corresponding to the position, velocity, and orientation of each agent sampled at exactly 10 Hz. As for the agents, various actors are present in the dataset: vehicles (both parked and moving), pedestrians, cyclists, scooters, and pets \cite{wilson2023argoverse}. However, since Cooperative Awareness Messages only consider vehicle-like agents, only passenger cars, motorcycles, and buses have been considered.

\looseness=-1
\nicepar{Dataset Statistics}. After the filtering process, passenger cars represent the vast majority of actors in the dataset, accounting for approximately 98\% of the total. Moreover, most of the scenarios contain more than 10 vehicles. Such a high number is crucial, as knowing the position of the neighbouring agents allows the neural network to infer admissible movements and other contextual cues. Lastly, by analyzing the speed distribution of the agents, many are either stationary or crawling.

\looseness=-1
\nicepar{Dataset Limitations}. Firstly, there is a lack of traffic information: for instance, no data about semaphores and road signs are present. Secondly, the scenarios present in Argoverse 2 only come from one country, the United States of America; thus, the road distribution can heavily differ from that found in other regions or continents. For instance, roundabouts are not as frequent in the US as they are in European cities. Lastly, a single driving style has been considered.

\subsection{CAM-based Dataset}
\begin{figure}[t]
    \centering
    \subfloat[Before interpolation]{\includegraphics[width=0.45\textwidth]   {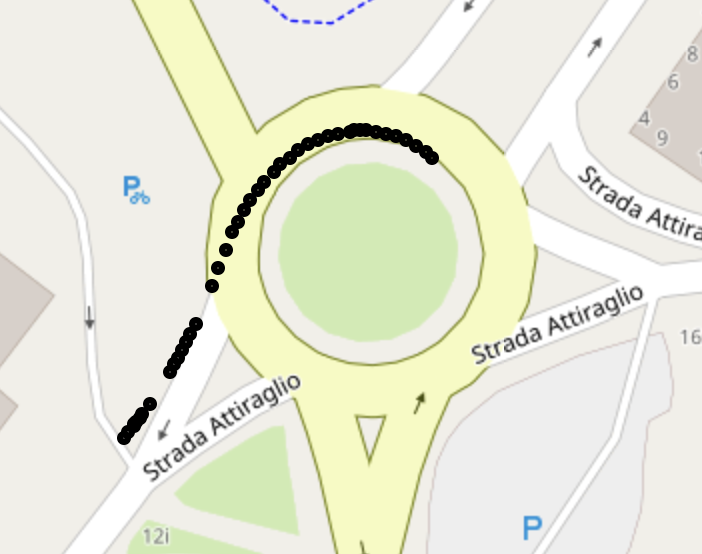}\label{fig:f1}}
    \hfill
    \subfloat[After interpolation]{\includegraphics[width=0.45\textwidth]
    {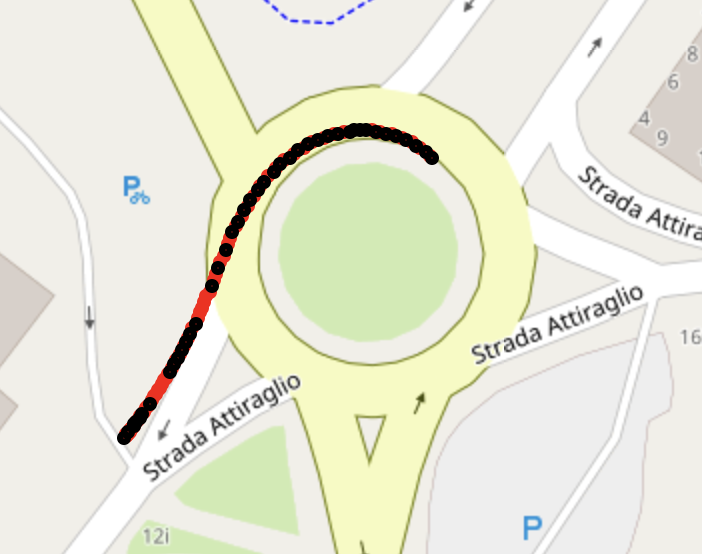}\label{fig:f2}}
    \caption{Illustration of the first interpolation performed. Original CAM data (black) and interpolated values (red) are shown.}\label{fig:dataset_adaptation}
\end{figure}

\looseness=-1
The second dataset has been constructed from Cooperative Awareness Messages gathered in Modena (Italy). Overall, 578 PCAP files were collected and processed to extract CAMs, which served to build scenarios similar to those of \cite{wilson2023argoverse}.

\looseness=-1
The preprocessing was performed as follows. First, data cleaning was carried out: if the same Cooperative Awareness Message was received multiple times by an RSU, only the first instance was retained. Moreover, all CAM data for which latitude, longitude, speed, or heading were missing were discarded and not used. To match the agents' features present in Argoverse 2, each vehicle's position is also represented in the UTM format. A first stage of interpolation is then performed to bring data at about 10Hz (see \cref{fig:dataset_adaptation}). In particular, data interpolation is carried out whenever the generation time between two consecutive CAMs of the same vehicle is below one second. At this point, 11-second-long scenarios can be created, and a second stage of interpolation is performed to bring the data to exactly 10Hz. Lastly, each scenario was manually analyzed to remove ambiguous scenes.

\looseness=-1
\nicepar{Dataset statistics}. After processing, $16,051$ scenarios were obtained and split into training and validation sets with an $80$–$20$ ratio. Approximately $98$\% of the scenarios contain only one agent, and none include more than three. Moreover, vehicle speed distribution shows notable differences from Argoverse 2, with fewer vehicles moving at near-zero speeds.

\looseness=-1
\nicepar{Dataset Limitations}. Similarly to Argoverse 2, no data about semaphores and road signs are present. Additionally, missing data may occur since, as of today, only a limited number of vehicles are technologically equipped to transmit CAMs reporting their status to others. Thus, the number of vehicles transmitting such data may differ from the actual number of vehicles present in the scenario.

\looseness=-1
\subsection{Metrics}
The metrics reported in this study are $\text{AvgMin}_k\text{ADE}$ (\textit{Average Displacement Error}), 
$\text{AvgMin}_k\text{FDE}$ (\textit{Final Displacement Error}), and $\text{AvgMR}_k$ (\textit{miss rate}). Here, $k$ denotes the number of predictions generated for each vehicle: if $k>1$, multiple predictions are produced and the one that minimizes the metric is selected for evaluation. In our experiments, the value of $k$ is set to both $1$ (single-path prediction) and $6$ (multi-path prediction). The latter protocol is used to assess the diversity and plausibility of trajectories produced by stochastic (multi-modal) predictors and is a standard evaluation practice in the trajectory-prediction literature \cite{gupta2018social, benaglia2024trajectory}.

\looseness=-1
We now present the definitions of ADE, FDE, and MR:

\begin{enumerate}
    \item \textit{Average Displacement Error} (ADE): the average $\ell_2$ distance between all ground-truth positions and their predicted counterparts:
    \tightEq
    \begin{equation}
     ADE=\frac{ \sum_{i=1}^N\sum_{t=T_{\operatorname{observ}}}^{T_{\operatorname{pred}}}{||\textbf{x}_i^t-\hat{\textbf{x}}_i^t||_2}}{N*(T_{\operatorname{pred}}\space-\space T_{\operatorname{observ}})}
    \end{equation}
    \item \textit{Final Displacement Error (FDE)}: the $\ell_2$ distance between the last ground truth position and the last predicted position.
    \tightEq
    \begin{equation}
        FDE=\frac{ \sum_{i=1}^N||\textbf{x}_i^{T_{pred}}-\hat{\textbf{x}}_i^{T_{pred}}||_2}{N}
    \end{equation}
    \item \textit{Miss Rate (MR)}: ratio of data that are not within 2.0 meters from the ground truth.
\end{enumerate}

\section{Proposed Model}
\label{sec_nn}
\begin{figure*}[!t]
    \centering
    \subfloat[Overview of \methname{}.]{\includegraphics[width=0.31\textwidth]   {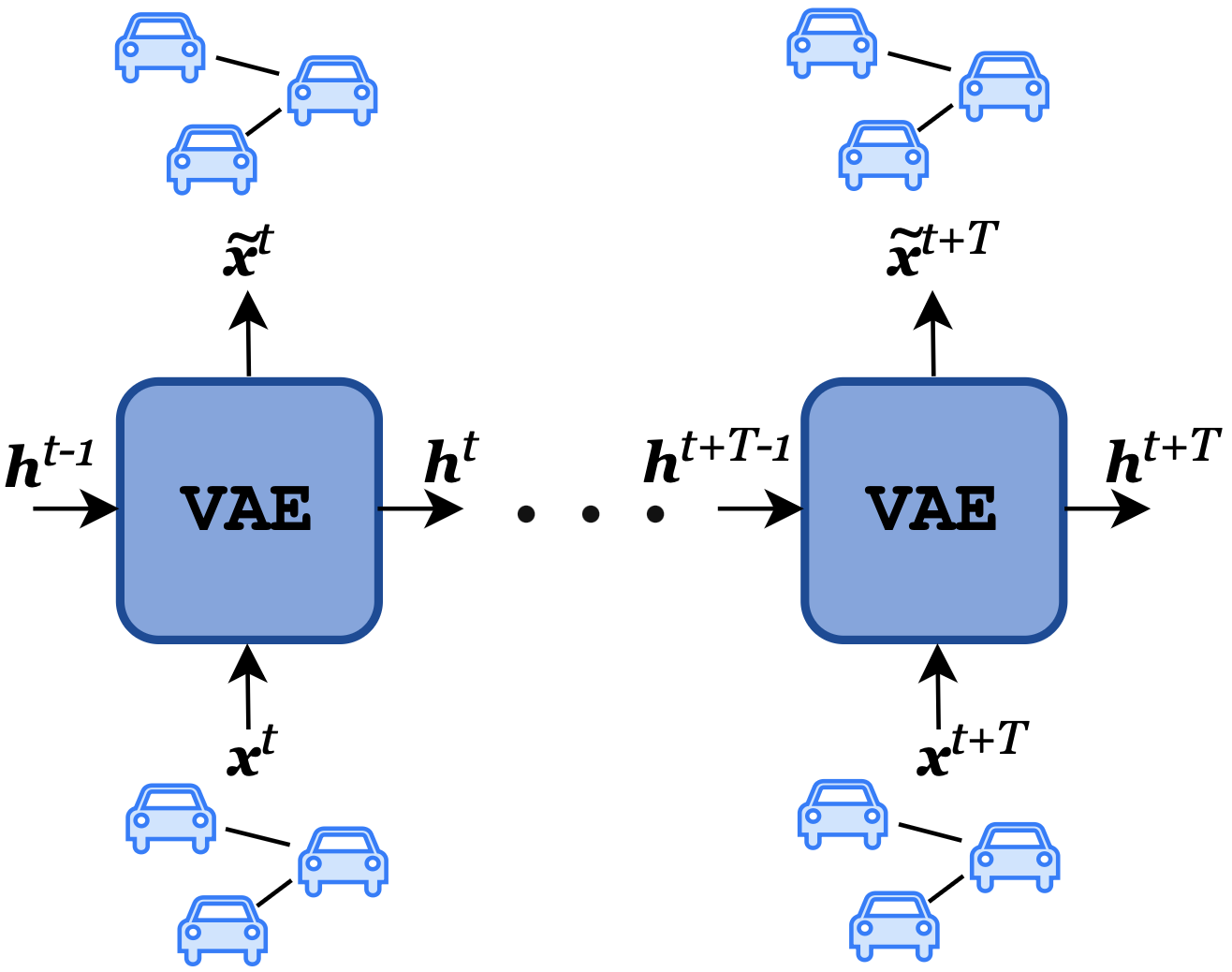}\label{fig:GraphTP}}
    \hspace{0.2cm}
    \subfloat[VAE structure at t-th time step.]{\includegraphics[width=0.27\textwidth]
    {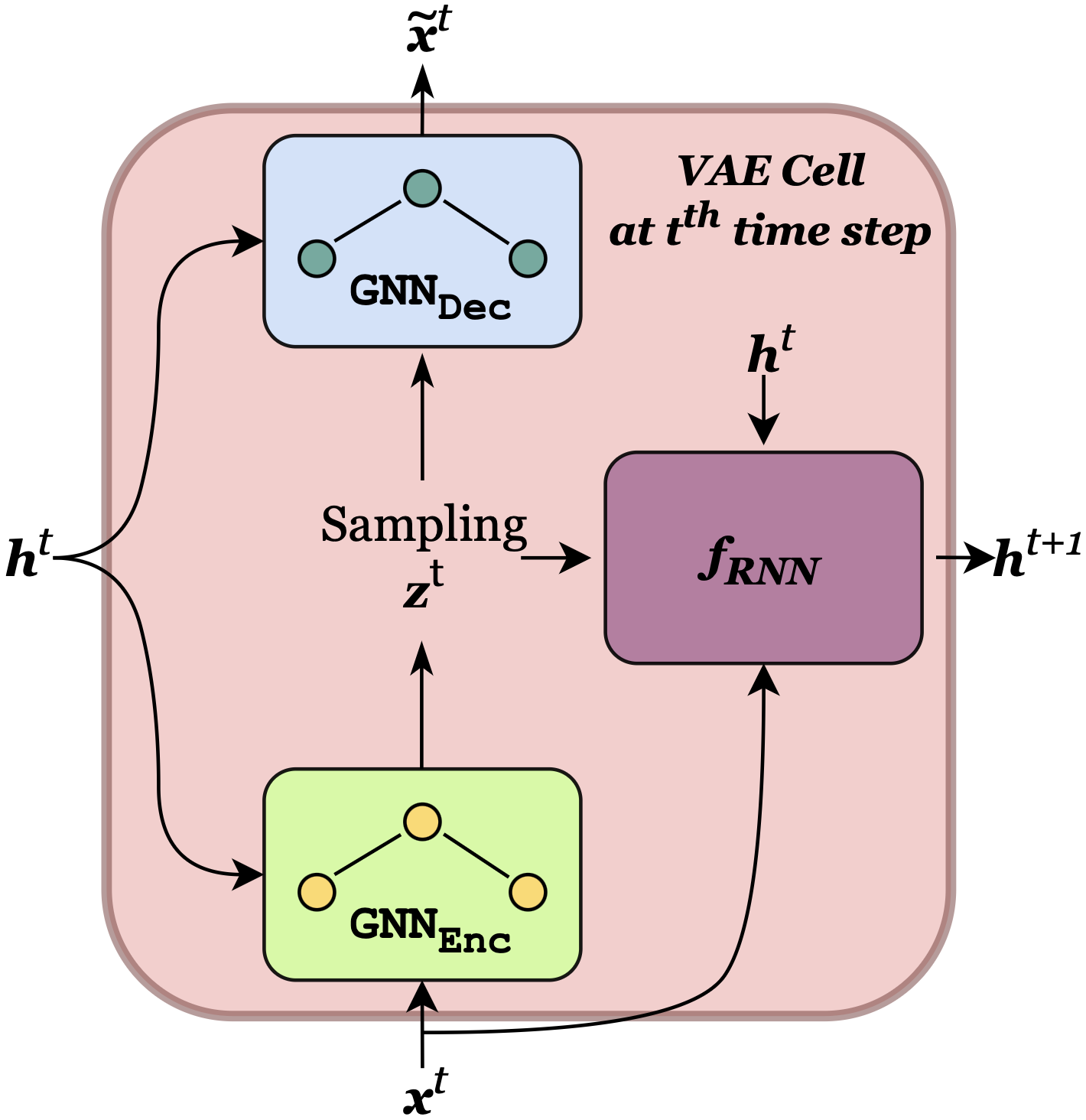}\label{fig:VAE_block}}
    \hspace{0.6cm}
    \subfloat[Encoder-decoder block.]{\includegraphics[width=0.20\textwidth]
    {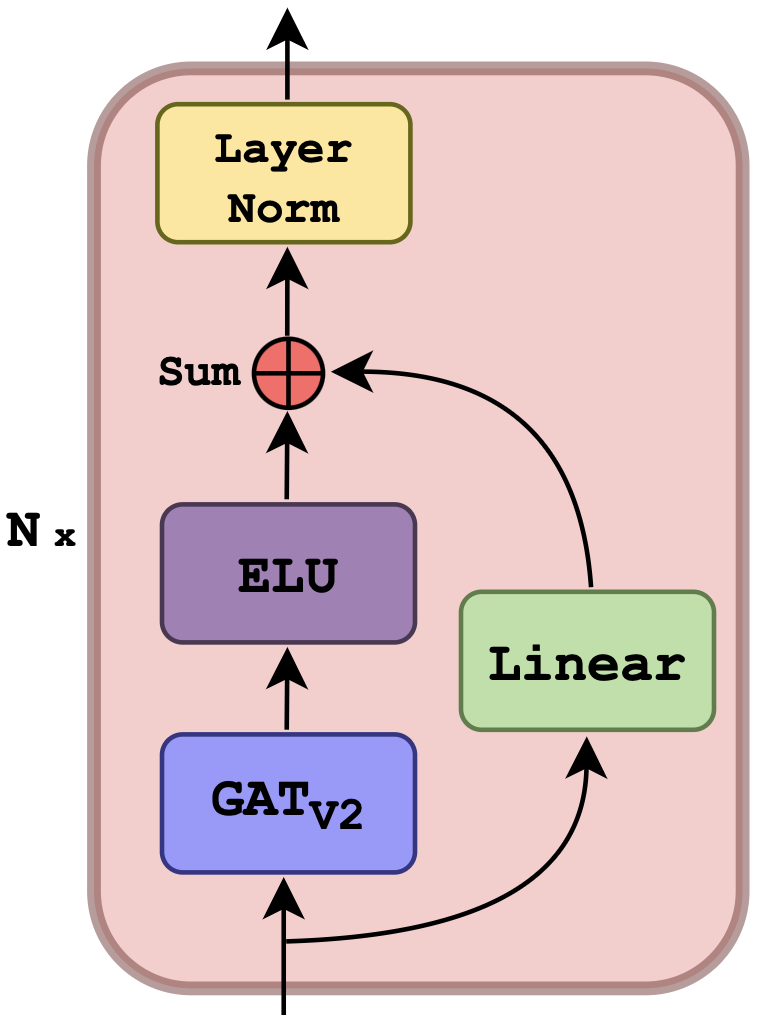}\label{fig:E_D_Block}}
    \caption{Brief description of the overall structure and main components of \methname{}.}
\end{figure*}

\looseness=-1
The neural network designed is called \extendedmethname{}, briefly \methname{}, which is an adaptation of \cite{YehCVPR2019} for the vehicular domain (see~\cref{fig:GraphTP,fig:VAE_block}).

\subsection{Architecture Overview}
\looseness=-1
Following the architecture of~\cite{YehCVPR2019}, \methname{} builds on the Variational Recurrent Neural Network introduced by~\cite{chung2015recurrent} to predict multiple plausible trajectories. The architecture is organized into three main building blocks -- encoder, decoder, and prior network -- and leverages Graph Neural Networks (GNNs) to model interactions among vehicles. The three aforementioned modules process and refine the following variables:
\begin{itemize}
    \item \textit{Observed variables} -- $x^t$: they represent the vehicles' information present in a given timestamp.
    \item \textit{Latent random variables} -- $z^t$: \qmarks{designed to capture the variations in the observed variables $x^t$} \cite{chung2015recurrent}. 
    \item \textit{Internal hidden state} -- $h^t$: variables that summarizes both the previous observed variables $x^{\leq t}$ and the stochastic choices $z^{\leq t}$ \cite{YehCVPR2019}. 
\end{itemize}

\looseness=-1
They are now describing the prior, decoder, and encoder distributions as well as the RNN update equation of the model:
\begin{align*}
p_{\theta}(z_t \,|\, x^{<t}, z^{<t}) &= \prod_k \mathcal{N}(z^t \,|\, \mu^t_{\operatorname{prior},k}, \sigma^t_{\operatorname{prior},k})
\ \texttt{\small (prior)} \\
p_{\theta}(x^t \,|\, x^{<t}, z^{\leq t}) &= \prod_k \mathcal{N}(x^t \,|\, \mu^t_{\operatorname{dec},k}, \sigma^t_{\operatorname{dec},k}) 
\ \texttt{\small (inference)} \\
q_\phi(z^t \,|\, x^{\leq t}, z^{<t}) &= \prod_k \mathcal{N}(z^t \,|\, \mu^t_{\operatorname{enc},k}, \sigma^t_{\operatorname{enc},k}) 
\ \texttt{\small (generate)} \\
h_k^t &= f_{\operatorname{RNN}}(\varphi^{x}(x^t_k), \varphi^{z}(z^t_k), h^{t-1}_k) \\
\textbf{where} \quad \quad \quad \quad & \\
\mu_{\operatorname{prior},1:K}^t, \sigma_{\operatorname{prior},1:K}^t 
&= \operatorname{GNN}_{\operatorname{prior}}(h^{t-1}_{1:K}) \\
\mu^t_{\operatorname{dec},1:K}, \sigma^t_{\operatorname{dec},1:K} 
&= \operatorname{GNN}_{\operatorname{dec}}([\varphi^{z}(z^t_{1:K}), h^{t-1}_{1:K}]) \\
\mu^t_{\operatorname{enc},1:K}, \sigma^t_{\operatorname{enc},1:K} 
&= \operatorname{GNN}_{\operatorname{enc}}([\varphi^{x}(x^t_{1:K}), h^t_{1:K}])
\end{align*}
\looseness=-1
In these equations, both \text{$\varphi^{\textbf{x}}$} and \text{$\varphi^{\textbf{z}}$} are linear layers, while \text{$\operatorname{GNN}_{\operatorname{prior}}$}, \text{$\operatorname{GNN}_{\operatorname{enc}}$}, \text{$\operatorname{GNN}_{\operatorname{dec}}$} represent the Graph Neural Networks (GNNs) employed in the prior, encoder, and decode, respectively. As for $\mathcal{N}(\boldsymbol{\cdot}| \mu,\sigma)$, it denotes a multivariate normal distribution \cite{YehCVPR2019}. In our approach, the prior network predicts a distribution over the latent variables at each time step based on the past hidden state, while the encoder infers the approximate posterior distribution of the latent variables given the current observation and past context. The decoder then generates the reconstruction of the current observation conditioned on both the latent variables and the recurrent hidden state. 

\looseness=-1
\methname{} is trained by maximizing the Evidence Lower BOund (ELBO). However, unlike \cite{chung2015recurrent} and \cite{YehCVPR2019}, we introduce a parameter $\beta$ to explicitly control, during training, the trade-off between reconstruction accuracy and adherence to the prior, following the formulation in \cite{higgins2017betavae}.

\begin{align}
    \begin{split} \label{eq:objective_function}
        \sum_{x \in \mathcal{D}} & \sum_t \mathbb{E}_{q_\phi(z^t \,|\, x^{\leq t}, z^{<t})}
        \Big[ \log p_\theta(x^t \,|\, x^{<t}, z^{\leq t}) \\ 
        &\quad - \beta \, D_{\operatorname{KL}}\!\big(q_\phi(z^t \,|\, x^{\leq t}, z^{<t}) \,\|\, p_\theta(z^t \,|\, x^{<t}, z^{<t})\big) \Big]
    \end{split}
\end{align} 
\subsection{Encoder, decoder, and prior blocks}
\label{sub_sec_blocks}
Drawing inspiration from the attention mechanism introduced in \cite{vaswani2017attention}, the foundational block used in both the encoder and decoder of \methname{} (see \cref{fig:E_D_Block}) consists of a GATv2 layer \cite{brody2021attentive}. GATv2 extends the original Graph Attention Network (GAT) \cite{velivckovic2017graph} by overcoming its static attention limitation. In our implementation, the message-passing step is followed by an Exponential Linear Unit (ELU), and the final output is concatenated with a linear projection of the graph layer input, thus forming a residual connection. Finally, layer normalization is performed.

\looseness=-1
Each graph layer inputs an adjacency matrix that captures agent relationships, providing an inductive bias on spatial locality and proximity during prediction. To this end, we explored three connection strategies: \textit{i)} All-to-All: every agent is connected to all others; \textit{ii)} K-Nearest Neighbors (KNN): each agent is connected to its $k$ closest neighbors, with $k$ as a hyperparameter; \textit{iii)} Distance-based: agents are connected if their distance is below a predefined threshold hyperparameter.

\section{Experiments}
\label{sec_experiments}
\looseness=-1
This chapter presents the experiments conducted on both datasets. Results on Argoverse 2 provide a benchmark for \methname{} and the competing methods, while experiments on the CAM-based dataset investigate the feasibility of using this data for vehicle trajectory prediction. 
\subsection{Argoverse 2 Motion Forecasting Dataset}
\label{subsec:argoverse_2}
The method proposed is compared to both context-free and context-aware models. We report the following baselines:
\begin{itemize}
    \item \textbf{Constant Velocity Model (CVM)} \cite{scholler2020constant} – A context-free approach that extrapolates future trajectories by assuming constant velocity and heading.  
    \item \textbf{LSTM} – A context-free recurrent model that predicts future positions solely from the agent’s past trajectory without leveraging interactions.  
    \item \textbf{VRNN} \cite{chung2015recurrent} – A context-free generative model that incorporates latent variables within a recurrent architecture to capture stochastic trajectory evolution.  
    \item \textbf{Forecast-MAE} \cite{cheng2023forecast} – A context-aware Transformer-based approach that models both temporal dynamics and agent interactions using masked autoencoders.  
\end{itemize}

\begin{table*}[t]
    \centering
    \renewcommand{\arraystretch}{1.0}
    \resizebox{\textwidth}{!}{%
    \begin{tabular}{lcccccccc}
    \toprule
    \textbf{Methods} & \textbf{Context} & $\text{AvgMin}_1\text{FDE}$ & $\text{AvgMin}_1\text{ADE}$ & $\text{Avg}_1\text{MR}$ & $\text{AvgMin}_6\text{FDE}$ & $\text{AvgMin}_6\text{ADE}$ & $\text{Avg}_6\text{MR}$  \\
    \midrule
    \textbf{CVM} \cite{scholler2020constant} & \ding{55} & $6.025$ & $2.326$ & $0.941$ & -- & -- & -- \\
    \textbf{LSTM} & \ding{55} & $\mathbf{5.217}$ & $\mathbf{2.005}$ & $\mathbf{0.459}$ & -- & -- & -- \\
    \textbf{VRNN} \cite{chung2015recurrent} & \ding{55} & $13.773$ & $6.852$ & $0.468$ & $5.892$ & $2.425$ & $\mathbf{0.444}$ \\ 
    \textbf{\methname{} (ours)} & \ding{55} & $7.779$ & $3.009$ & $0.545$ & $\mathbf{3.887}$ & $\mathbf{1.663}$ & $0.524$ \\
    \midrule
    \textbf{Forecast-MAE} \cite{cheng2023forecast} & \ding{51} & $4.846$ & $1.833$ & $0.438$ & $1.680$ & $0.739$ & $0.203$ \\
    \bottomrule
    \end{tabular}
    }
    \caption{Results obtained on the validation set of Argoverse 2 Motion Forecasting Dataset.}
    \label{tab:AV2_results}
\end{table*}

\looseness=-1
As previously described, \methname{} does not incorporate any context information; therefore, the comparison with Forecast-MAE highlights the extent to which the absence of road-related information affects the final results. All models have been retrained from scratch, and results are reported in \cref{tab:AV2_results}.

\looseness=-1
\nicepar{Setup.} The model takes as input the relative position and velocity, while heading information is handled during preprocessing. Both the encoder and decoder block present two graph-based neural blocks, discussed in \cref{sub_sec_blocks}; as for the prior, instead, only one is used. All GATv2 layers contain four attention heads. The intermediate and graph representations share the same dimensionality ($64$), while the latent dimensionality is set to $16$. The model was trained for $60$ epochs using the Adam optimizer~\cite{kingma2015adam} with mini-batches of size $128$, weight decay equal to \text{$1\times10^{-4}$}, and an initial learning rate of $2\times10^{-4}$, decayed to $1\times10^{-6}$ via a cosine-annealing scheduler. Lastly, the \text{$\beta$} parameter in \cref{eq:objective_function} is initially set to $0$ and it linearly increases to $1.0$ in $15$ epochs as a form of warm-up.

\looseness=-1
\nicepar{Results}. From \cref{tab:AV2_results}, \methname{} outperforms VRNN in the multi-path setting ($k=6$), indicating that explicitly modeling inter-agent interactions provides a tangible benefit. In contrast, in the single-path setting ($k=1$), a simpler deterministic model such as the LSTM attains better scores across all metrics -- this scenario tends to favor models trained to predict a single trajectory deterministically. In both settings, the best overall performance is achieved by Forecast-MAE (context-aware), which leverages HD-maps and contextual information, underscoring their importance for reliable trajectory prediction. 

\begin{figure}[t]
    \centering
    \subfloat[]{\includegraphics[width=0.49\textwidth] {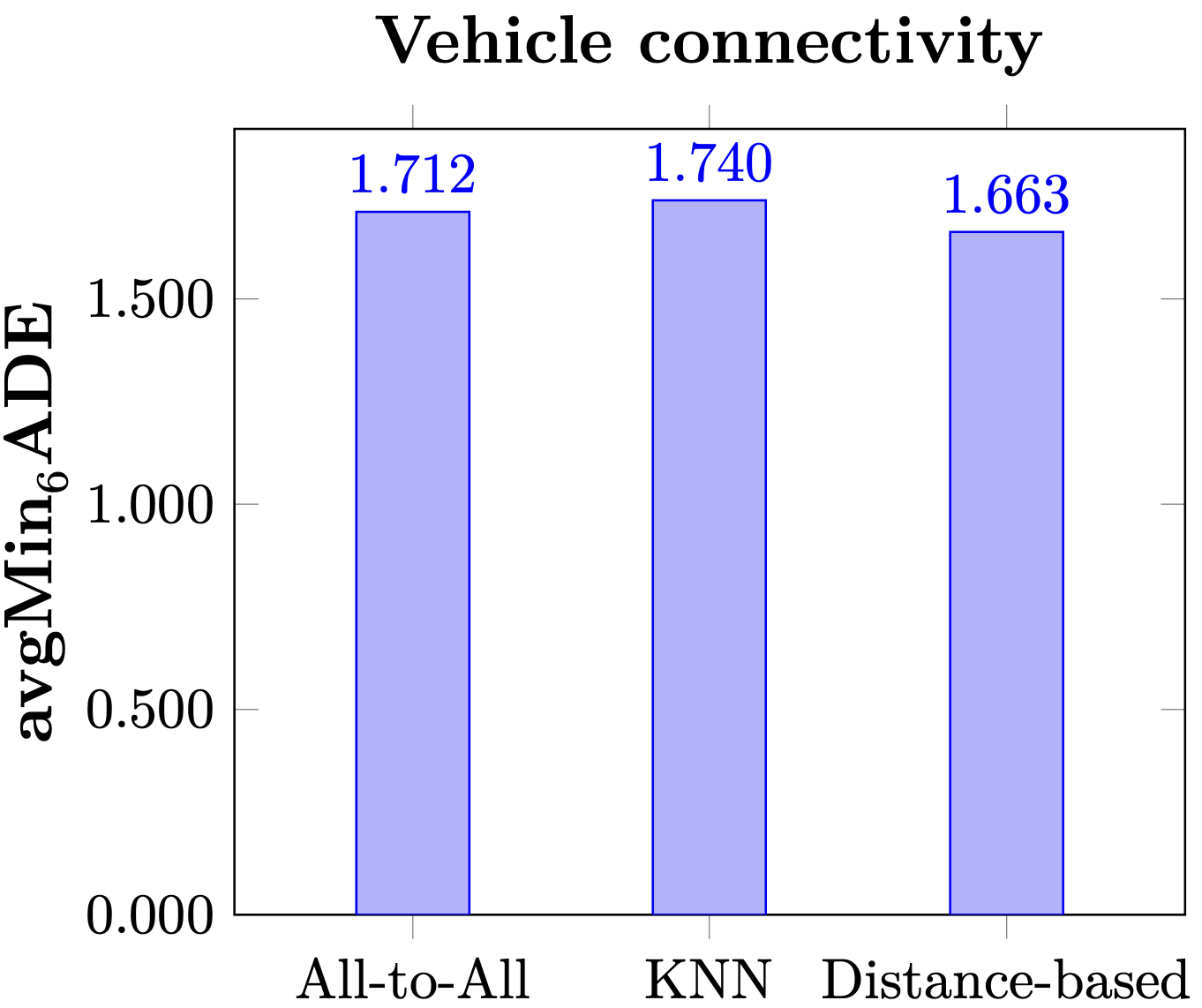}\label{fig:f1_ablation}}
    \hfill
    \subfloat[]{\includegraphics[width=0.475\textwidth]
{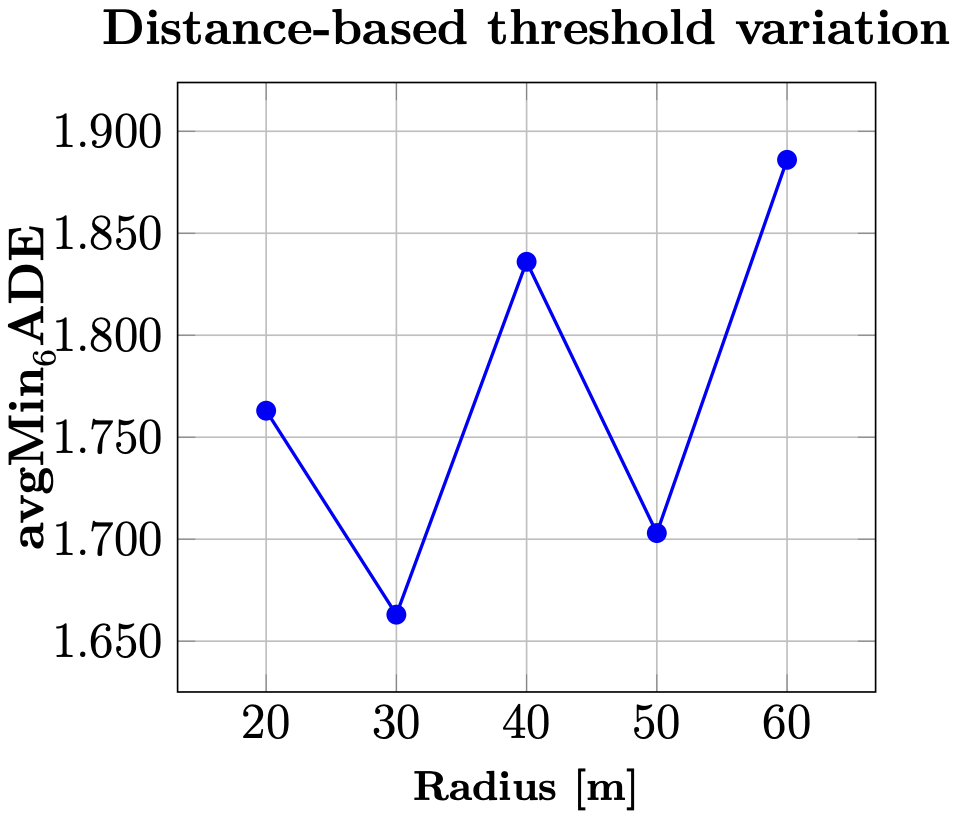}\label{fig:f2_ablation}}
    \caption{Analysis on the variation of the vehicle connectivities (a) and distance thresholds (b) }\label{fig:ablation_study}   
\end{figure}

\looseness=-1
\nicepar{Ablation study}. In the ablation study, we use $\text{AvgMin}_6\text{ADE}$ as the primary comparison metric. First, we compare the three agent-connectivity strategies described in \cref{sub_sec_blocks}. In \cref{fig:f1_ablation}, distance-based connectivity achieves the best results when the threshold is set to 30 meters. This outcome might be due to the fact that, when using all-to-all or KNN connectivity, the neural network must learn, during training, to ignore distant vehicles as their information content is generally minimal. Another analysis concerns the choice of the distance threshold for distance-based connectivity; as previously discussed, $30$m is optimal. As shown in \cref{fig:f2_ablation}, larger thresholds include more distant vehicles, requiring the network to down-weight or ignore them. Instead, when the threshold is reduced, \methname{} cannot exploit the information of neighbouring vehicles, thereby affecting the final results. Lastly, the insertion of residual connections in graph layers has been investigated and shown to boost the performance, reducing $\text{AvgMin}_6\text{ADE}$ from $2.264$ to $1.663$. 

\addtolength{\textfloatsep}{-16pt}

\looseness=-1
\subsection{CAM-based Dataset}
On the CAM-based dataset, we restrict comparisons to context-free models -- CVM, LSTM, and VRNN -- because only CAMs are available and no HD-maps are provided. Consequently, context-aware baselines cannot be evaluated.

\looseness=-1
To assess how \methname{} and competing methods generalize to novel scenarios, we consider two evaluation regimes: \textbf{zero-shot} and \textbf{fine-tuning}. In the first setting, we evaluate the best-performing checkpoints trained on Argoverse 2 directly on the CAM-based dataset, without further training. In the second setting, all models -- except CVM -- are fine-tuned on the CAM-based dataset starting from the same checkpoints used for the zero-shot evaluation, then the results are reported. This protocol quantifies both out-of-distribution generalization and the gains achievable when adapting to CAM data.

\looseness=-1
\nicepar{Setup}. The model is trained end-to-end for $60$ epochs. The learning rate starts at $5\times10^{-4}$ and decays to $1\times10^{-6}$ by epoch $60$ via cosine annealing. The $\beta$ coefficient is linearly warmed up from $0$ to $0.1$ over the first $15$ epochs and then kept at $0.1$. All other hyperparameters follow \cref{subsec:argoverse_2}.

\begin{figure}[!t]
    \centering
    \subfloat[\methname{}]{\includegraphics[width=0.38\textwidth]   {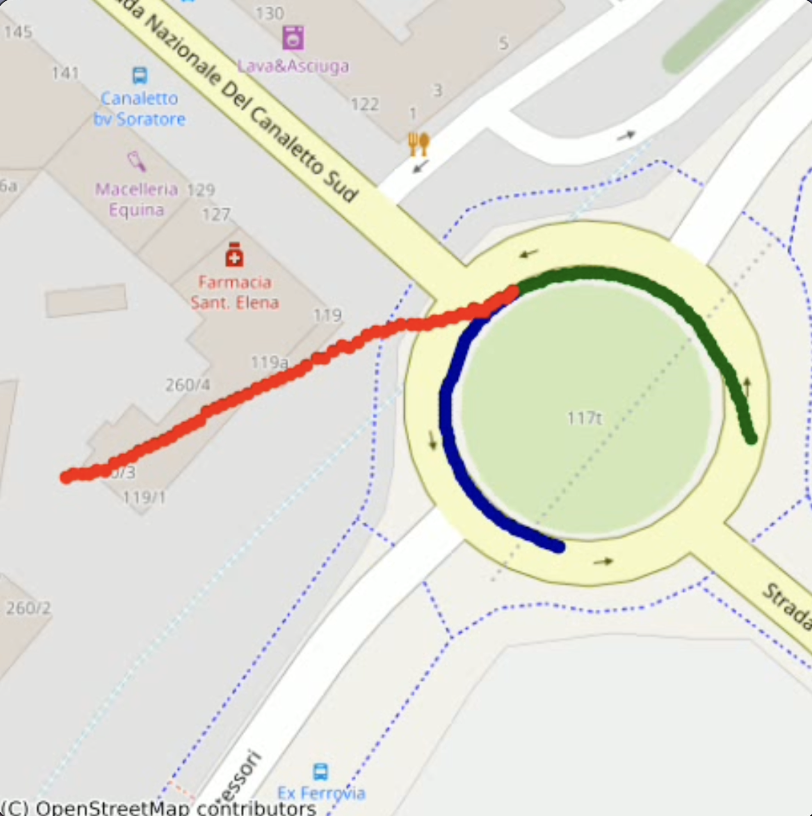}\label{fig:f1_bad_predictions}}
    \hspace{0.7cm}
    \subfloat[LSTM]{\includegraphics[width=0.38\textwidth]
    {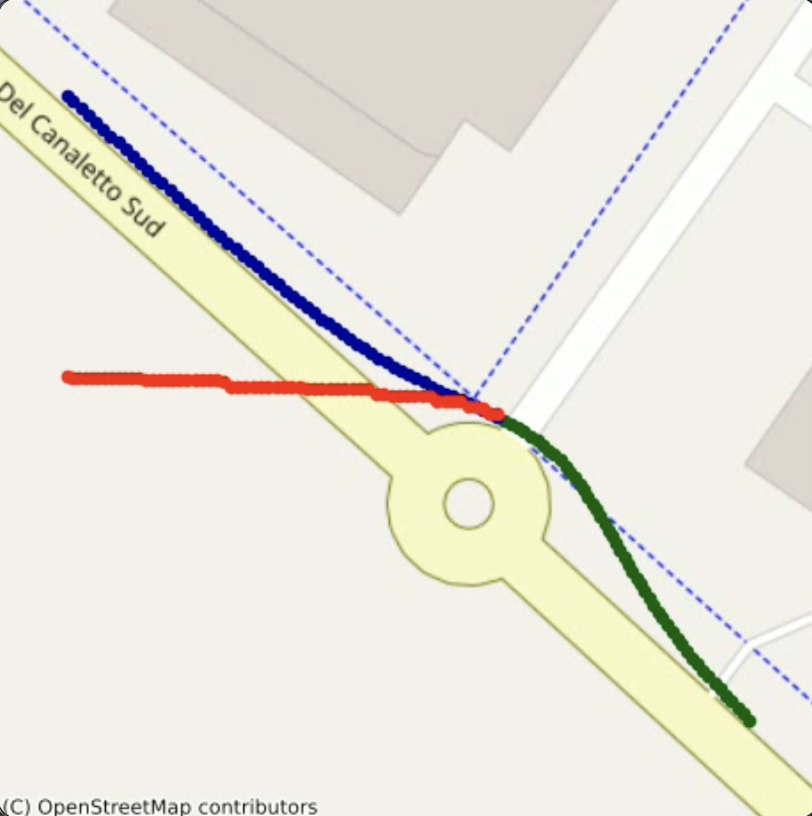}\label{fig:f2_bad_predictions}}
    \caption{Illustration of incorrect predictions. Observations (green), blue (ground truth), and predictions (red) are shown.}\label{fig:bad_predictions}
\end{figure}

\looseness=-1
\nicepar{Results}. As shown in \cref{tab:custom_results}, in the zero-shot setting, the simple CVM achieves the best results, indicating limited transfer from Argoverse 2 to our CAM-based dataset. After fine-tuning, performance improves: Our approach surpasses CVM, indicating that it captures more complex interaction patterns and thereby justifies the use of data-driven learning techniques for CAM-based trajectory prediction. However, performance on the CAM-based dataset remains markedly worse than on Argoverse 2 (\cref{tab:AV2_results}). We attribute this gap to a pronounced distribution shift and, more importantly, to greater trajectory complexity in our data (e.g., more intricate routes and maneuvers). Moreover, the absence of contextual priors -- such as HD-maps -- prevents the disambiguation of lane geometry and affordances. In addition, missing data and the small number of simultaneously observed agents limit interaction cues, further degrading prediction quality. In \cref{fig:bad_predictions}, two visual examples of incorrect predictions are reported.

\begin{table*}[t]
    \centering
    \renewcommand{\arraystretch}{1.0}
    \resizebox{\textwidth}{!}{%
    \begin{tabular}{llcccccccc}
    \toprule
    & \textbf{Methods} & $\text{AvgMin}_1\text{FDE}$ & $\text{AvgMin}_1\text{ADE}$ & $\text{Avg}_1\text{MR}$ & $\text{AvgMin}_6\text{FDE}$ & $\text{AvgMin}_6\text{ADE}$ & $\text{Avg}_6\text{MR}$  \\
    \midrule
    \multirow{4}{*}{\textbf{\textit{Zero-shot}}} &
    \textbf{CVM} \cite{scholler2020constant} & $\mathbf{16.134}$ & $\mathbf{7.557}$ & $\mathbf{0.886}$ & -- & -- & -- \\ &
    \textbf{LSTM} & $17.714$ & $7.981$ & $0.946$ & -- & -- & -- \\ &
    \textbf{VRNN} \cite{chung2015recurrent} & $47.075$ & $24.087$ & $0.889$ & $23.486$ & $11.113$ & $\mathbf{0.882}$ \\ &
    \textbf{\methname{} (ours)} & $36.868$ & $17.376$ & $0.937$ & $\mathbf{19.111}$ & $\mathbf{9.538}$ & $0.924$ \\
    \midrule
    \multirow{4}{*}{\textbf{\textit{Finetuning}}} &
    \textbf{LSTM} & $\mathbf{10.291}$ & $\mathbf{4.196}$ & $\mathbf{0.824}$ & -- & -- & -- \\ &
    \textbf{VRNN} \cite{chung2015recurrent} & $42.044$ & $22.026$ & $0.994$ & $19.009$ & $9.871$ & $0.982$ \\ &
    \textbf{\methname{} (ours)} & $31.669$ & $13.653$ & $0.984$ & $\mathbf{14.562}$ & $\mathbf{7.362}$ & $\mathbf{0.970}$ \\
    \bottomrule
    \end{tabular}
    }
    \caption{Results obtained on the validation set of the CAM-based dataset.}
    \label{tab:custom_results}
\end{table*} 

\section{Conclusions and Future Works}
\label{sec_conclusions}
\looseness=-1
In this paper, we studied the use of CAMs for predicting future vehicle states with modern neural networks. We detailed a processing pipeline that converts raw CAMs into benchmarking scenarios comparable to popular datasets, such as Argoverse 2. We highlighted various limitations: for instance, as of today, not enough vehicles are sufficiently technologically advanced to transmit such data, and thus, scenarios do not include all the vehicles really present. This last constraint has shown a strong negative impact on the results of the CAM-based dataset, especially because no context had been used.

\looseness=-1
Regarding future works, we plan to incorporate context information, which has proved beneficial on Argoverse~2, while exploring alternatives to strictly map-centric pipelines. When HD-maps are unavailable, contextual cues -- such as inter-vehicle distance estimates~\cite{panariello2025monocular} and scene appearance signals~\cite{mancusi2023trackflow} -- could be inferred from onboard vision sensors and fused into the model to improve reliability.

\bibliographystyle{IEEEtran}
\bibliography{citations}

\begin{thebibliography}{10}
\providecommand{\url}[1]{#1}
\csname url@samestyle\endcsname
\providecommand{\newblock}{\relax}
\providecommand{\bibinfo}[2]{#2}
\providecommand{\BIBentrySTDinterwordspacing}{\spaceskip=0pt\relax}
\providecommand{\BIBentryALTinterwordstretchfactor}{4}
\providecommand{\BIBentryALTinterwordspacing}{\spaceskip=\fontdimen2\font plus
\BIBentryALTinterwordstretchfactor\fontdimen3\font minus \fontdimen4\font\relax}
\providecommand{\BIBforeignlanguage}[2]{{%
\expandafter\ifx\csname l@#1\endcsname\relax
\typeout{** WARNING: IEEEtran.bst: No hyphenation pattern has been}%
\typeout{** loaded for the language `#1'. Using the pattern for}%
\typeout{** the default language instead.}%
\else
\language=\csname l@#1\endcsname
\fi
#2}}
\providecommand{\BIBdecl}{\relax}
\BIBdecl

\bibitem{Survey_autonomous_driving}
Y.~Huang \emph{et~al.}, ``A survey on trajectory-prediction methods for autonomous driving,'' \emph{IEEE Transactions on Intelligent Vehicles}, vol.~7, no.~3, pp. 652--674, 2022.

\bibitem{ETSI-v141}
ETSI, ``{ETSI-EN 302 637-2 v1.4.1. Intelligent Transport Systems (ITS); Specification of Cooperative Awareness Basic Service.}'' European Telecommunications Standards Institute, Tech. Rep., 2019.

\bibitem{chang2019argoverse}
M.-F. Chang \emph{et~al.}, ``Argoverse: 3d tracking and forecasting with rich maps,'' in \emph{Proceedings of the IEEE/CVF conference on computer vision and pattern recognition}, 2019, pp. 8748--8757.

\bibitem{wilson2023argoverse}
B.~Wilson \emph{et~al.}, ``Argoverse 2: Next generation datasets for self-driving perception and forecasting,'' \emph{arXiv preprint arXiv:2301.00493}, 2023.

\bibitem{interactiondataset}
W.~Zhan \emph{et~al.}, ``{INTERACTION} {Dataset}: {An} {INTERnational}, {Adversarial} and {Cooperative} {moTION} {Dataset} in {Interactive} {Driving} {Scenarios} with {Semantic} {Maps},'' \emph{arXiv:1910.03088 [cs, eess]}, Sep. 2019.

\bibitem{ettinger2021large}
S.~Ettinger \emph{et~al.}, ``{Large scale interactive motion forecasting for autonomous driving: The Waymo Open Motion Dataset},'' in \emph{Proceedings of the IEEE/CVF international conference on computer vision}, 2021, pp. 9710--9719.

\bibitem{houston2021one}
J.~Houston \emph{et~al.}, ``{One thousand and one hours: Self-driving motion prediction dataset},'' in \emph{Conference on Robot Learning}.\hskip 1em plus 0.5em minus 0.4em\relax PMLR, 2021, pp. 409--418.

\bibitem{caesar2020nuscenes}
H.~Caesar \emph{et~al.}, ``{nuscenes: A multimodal dataset for autonomous driving},'' in \emph{Proceedings of the IEEE/CVF conference CVPR}, 2020, pp. 11\,621--11\,631.

\bibitem{yu2023v2x}
H.~Yu \emph{et~al.}, ``{V2X-seq: A large-scale sequential dataset for vehicle-infrastructure cooperative perception and forecasting},'' in \emph{Proceedings of the IEEE/CVF Conference CVPR}, 2023, pp. 5486--5495.

\bibitem{yu2022dair}
------, ``{Dair-v2x: A large-scale dataset for vehicle-infrastructure cooperative 3d object detection},'' in \emph{Proceedings of the IEEE/CVF conference CVPR}, 2022, pp. 21\,361--21\,370.

\bibitem{scholler2020constant}
C.~Sch{\"o}ller \emph{et~al.}, ``{What the constant velocity model can teach us about pedestrian motion prediction},'' \emph{IEEE Robotics and Automation Letters}, vol.~5, no.~2, pp. 1696--1703, 2020.

\bibitem{monti2022many}
A.~Monti \emph{et~al.}, ``How many observations are enough? knowledge distillation for trajectory forecasting,'' in \emph{Proceedings of the IEEE/CVF Conference CVPR}, 2022, pp. 6553--6562.

\bibitem{liang2020learning}
M.~Liang \emph{et~al.}, ``{Learning lane graph representations for motion forecasting},'' in \emph{European Conference on Computer Vision}.\hskip 1em plus 0.5em minus 0.4em\relax Springer, 2020, pp. 541--556.

\bibitem{kipf2017semisupervisedclassificationgraphconvolutional}
T.~N. Kipf and M.~Welling, ``{Semi-Supervised Classification with Graph Convolutional Networks},'' 2017.

\bibitem{cheng2023forecast}
J.~Cheng, X.~Mei, and M.~Liu, ``{Forecast-MAE}: Self-supervised pre-training for motion forecasting with masked autoencoders,'' \emph{Proceedings of the IEEE/CVF International Conference on Computer Vision}, 2023.

\bibitem{he2022masked}
K.~He \emph{et~al.}, ``{Masked autoencoders are scalable vision learners},'' in \emph{Proceedings of the IEEE/CVF conference CVPR}, 2022, pp. 16\,000--16\,009.

\bibitem{gupta2018social}
A.~Gupta \emph{et~al.}, ``Social gan: Socially acceptable trajectories with generative adversarial networks,'' in \emph{Proceedings of the IEEE conference on computer vision and pattern recognition}, 2018, pp. 2255--2264.

\bibitem{benaglia2024trajectory}
R.~Benaglia \emph{et~al.}, ``Trajectory forecasting through low-rank adaptation of discrete latent codes,'' in \emph{International Conference on Pattern Recognition}.\hskip 1em plus 0.5em minus 0.4em\relax Springer, 2024, pp. 236--251.

\bibitem{YehCVPR2019}
R.~A. Yeh \emph{et~al.}, ``Diverse generation for multi-agent sports games,'' in \emph{Proc. CVPR}, 2019.

\bibitem{chung2015recurrent}
J.~Chung \emph{et~al.}, ``A recurrent latent variable model for sequential data,'' \emph{Advances in neural information processing systems}, vol.~28, 2015.

\bibitem{higgins2017betavae}
I.~Higgins \emph{et~al.}, ``beta-vae: Learning basic visual concepts with a constrained variational framework,'' in \emph{International conference on learning representations}, 2017.

\bibitem{vaswani2017attention}
A.~Vaswani \emph{et~al.}, ``Attention is all you need,'' \emph{Advances in neural information processing systems}, vol.~30, 2017.

\bibitem{brody2021attentive}
S.~Brody, U.~Alon, and E.~Yahav, ``How attentive are graph attention networks?'' \emph{arXiv preprint arXiv:2105.14491}, 2021.

\bibitem{velivckovic2017graph}
P.~Veli{\v{c}}kovi{\'c} \emph{et~al.}, ``Graph attention networks,'' \emph{arXiv preprint arXiv:1710.10903}, 2017.

\bibitem{kingma2015adam}
D.~P. Kingma and J.~Ba, ``Adam: A method for stochastic optimization,'' in \emph{International Conference on Learning Representations (ICLR)}, 2015.

\bibitem{panariello2025monocular}
A.~Panariello \emph{et~al.}, ``Monocular per-object distance estimation with masked object modeling,'' \emph{Computer Vision and Image Understanding}, vol. 253, p. 104303, 2025.

\bibitem{mancusi2023trackflow}
G.~Mancusi \emph{et~al.}, ``Trackflow: Multi-object tracking with normalizing flows,'' in \emph{Proceedings of the IEEE/CVF International Conference on Computer Vision}, 2023, pp. 9531--9543.

\end{thebibliography}

\end{document}